\documentclass[10pt,twocolumn,letterpaper]{article}

\usepackage{iccv}              

\definecolor{iccvblue}{rgb}{0.21,0.49,0.74}
\usepackage[pagebackref,breaklinks,colorlinks,allcolors=iccvblue]{hyperref}
\usepackage{booktabs}
\usepackage{makecell}

\def\paperID{1851} 
\def\confName{ICCV}
\def\confYear{2025}

\title{From Imitation to Innovation: The Emergence of AI's Unique Artistic Styles and the Challenge of Copyright Protection}

\author{
Zexi Jia\textsuperscript{1},
Chuanwei Huang\textsuperscript{2},
Yeshuang Zhu\textsuperscript{1},
Hongyan Fei\textsuperscript{2},
Ying Deng\textsuperscript{1},
Zhiqiang Yuan\textsuperscript{1},
\\
Jiapei Zhang\textsuperscript{1},
Jinchao Zhang\textsuperscript{1*},
Jie Zhou\textsuperscript{1}
\\
\textsuperscript{1}WeChat AI, Tencent Inc, China\\
\textsuperscript{2}Institute for Artificial Intelligence, Peking University
}

\begin{document}
\maketitle
\begin{abstract} 
Current legal frameworks consider AI-generated works eligible for copyright protection when they meet originality requirements and involve substantial human intellectual input. However, systematic legal standards and reliable evaluation methods for AI art copyrights are lacking. Through comprehensive analysis of legal precedents, we establish three essential criteria for determining distinctive artistic style: stylistic consistency, creative uniqueness, and expressive accuracy. To address these challenges, we introduce ArtBulb, an interpretable and quantifiable framework for AI art copyright judgment that combines a novel style description-based multimodal clustering method with multimodal large language models (MLLMs). We also present AICD, the first benchmark dataset for AI art copyright annotated by artists and legal experts. Experimental results demonstrate that ArtBulb outperforms existing models in both quantitative and qualitative evaluations. Our work aims to bridge the gap between the legal and technological communities and bring greater attention to the societal issue of AI art copyrights. 
\end{abstract}    
\renewcommand{\thefootnote}{}

\footnote{* Corresponding Author. This work has been accepted by ICCV 2025.}

\section{Introduction} \label{sec:intro}

\begin{figure*}[h!]
\begin{center}
\includegraphics[width=0.9\linewidth]{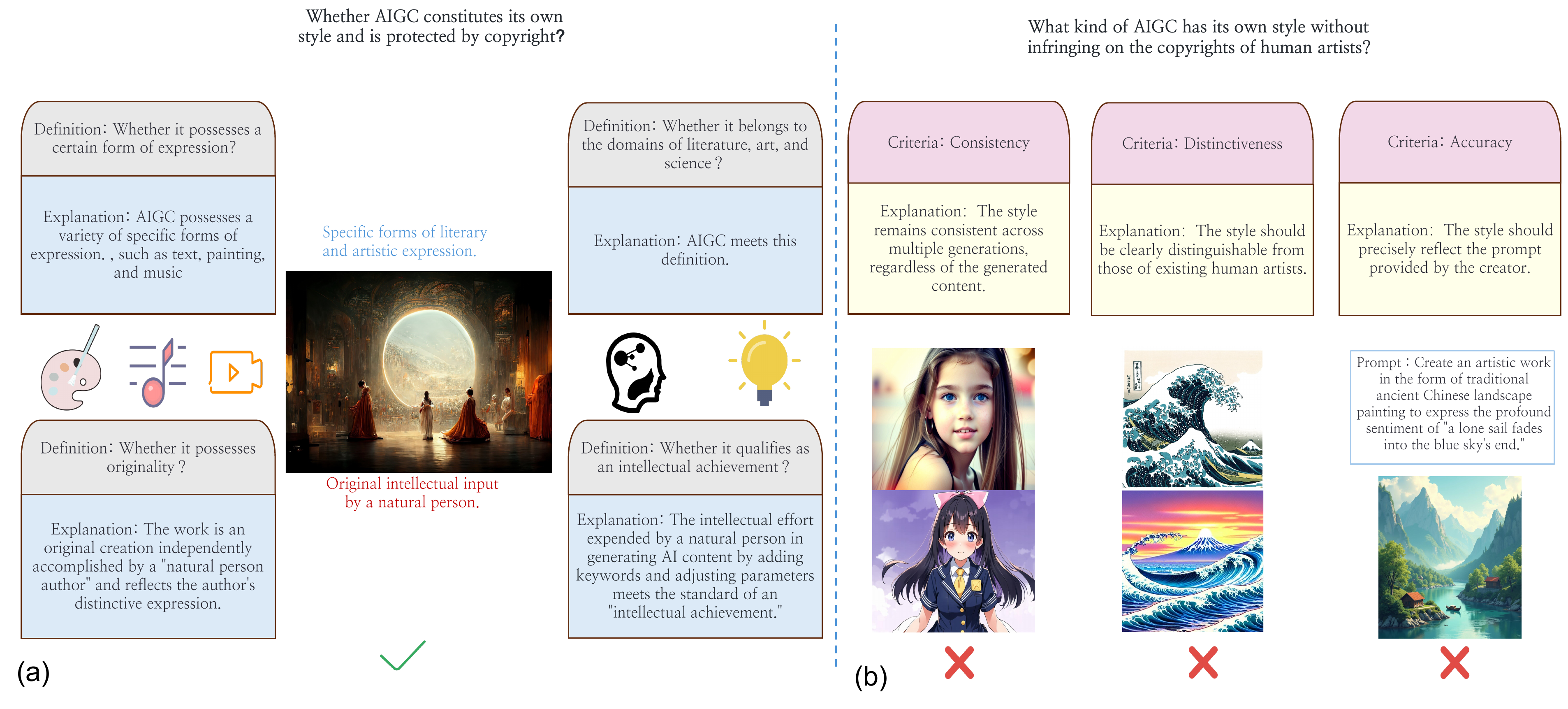}
\end{center}
   \caption{(a) Current laws in various countries determine whether AI-generated images qualify as protected works based on several criteria: they must belong to the realm of literature, art, or science; have a tangible form; exhibit originality; and originate from intellectual labor. (b) Based on these standards, we propose that protected AI works must meet three essential criteria: consistency, accuracy, and uniqueness.
}
\label{fig:fig1}
\end{figure*}

Artificial Intelligence Generated Content (AIGC) has transformed the creative landscape in recent years, sparking significant legal and ethical debates. A landmark ruling by the Beijing Court on November 27, 2023 \cite{he2024china}, establishes that AI-generated works meeting originality requirements and reflecting unique human intellectual contributions are entitled to the same copyright protection as human-created works. While this decision opens the door for recognizing and commercializing AI-generated content, granting it equal legal status with human creations, it also raises critical questions about style recognition and copyright protection in AIGC, particularly in defining unique AI styles and safeguarding artists' rights.

Addressing these issues presents multidisciplinary challenges across law, art, and technology, with far reaching economic, cultural, and ethical implications. A key criterion for determining copyright infringement is assessing substantial similarity between the original and the allegedly infringing work \cite{goldstein2005goldstein}. As shown in Figure~\ref{fig:fig1}(a), current laws require AI-generated images to belong to literature, art, or science; have a tangible form; exhibit originality; and originate from intellectual labor to qualify for copyright protection. 

To tackle these challenges, we propose an intuitive, automated, and legally sound method to determine whether AI-generated works can hold independent copyrights without infringing on human creators' rights. Our solution aligns with global copyright laws and is accessible to nontechnical stakeholders like judges, artists, and publishers. Based on legal precedents \cite{regulations2020tencent,wen2024copyright,kahveci2023attribution}, we establish that for AI-generated images to obtain copyright protection, they should meet the following criteria (see Figure~\ref{fig:fig1}(b)):

\begin{itemize}
    \item \textbf{Consistency}: The style remains consistent across multiple generations, regardless of content.
    \item \textbf{Uniqueness}: The style is significantly different from existing human artists' styles.
    \item \textbf{Accuracy}: The style accurately reflects the prompts provided by the creator.
\end{itemize}

We address style recognition as a machine learning problem by introducing a novel method named Description Guided Clustering (DGC). This method enhances the accuracy of artistic style clustering. It extracts key features representing different artistic styles from a large corpus of human artists' works. AI-generated works that form a unique style, usually originating from specific prompts and fine - tuned generative models, display consistent styles recognizable by our clustering model. Importantly, AI artists' styles should be clearly different from those of human artists, ensuring that AI-generated content is both coherent and unique. Also, AI artworks should accurately reflect the prompts, precisely depicting the visual elements, layout, and composition described, thus embodying the creator's personalized expression to meet the originality requirement.

We develop an interpretable copyright protection system, ArtBulb, based on Vision Language Models (VLMs). ArtBulb helps AI creators determine whether their works exhibit their own style, assess potential infringement, and evaluate copyright status.

To support our approach, we compile a dataset of artists' works showing unique styles, which serves as a baseline for style comparison. We also collect AI-generated works certified as having their own copyrights, as well as those considered infringing or lacking copyright protection. Together, these form the AI Copyright Dataset (AICD), the first dataset dedicated to addressing the copyright challenges of AI-generated works.

Our research bridges the gap between legal and technological communities and aims to draw more attention to AI art copyright issues. By providing a practical tool for navigating the complexities of AIGC in the context of intellectual property rights, we contribute to the ongoing discussion on the legal recognition and protection of AI-generated creative works.

In summary, our contributions are as follows:

\begin{itemize}
    \item We propose three core criteria for the copyright protection of AI artworks, transform the problem into a clustering task, and offer a practical solution.
    \item We develop the AI Copyright Dataset (AICD), the first dataset dedicated to AI-artwork copyright protection.
    \item We introduce a new clustering method based on style descriptions, Description-Guided Clustering (DGC), which enhances clustering accuracy.
    \item We build an explainable copyright framework, ArtBulb, validated by experts for interpretive judgments.
\end{itemize}
\section{Related Work}
\label{sec:formatting}

\begin{figure*}[h!]
\begin{center}
\includegraphics[width=0.9\linewidth]{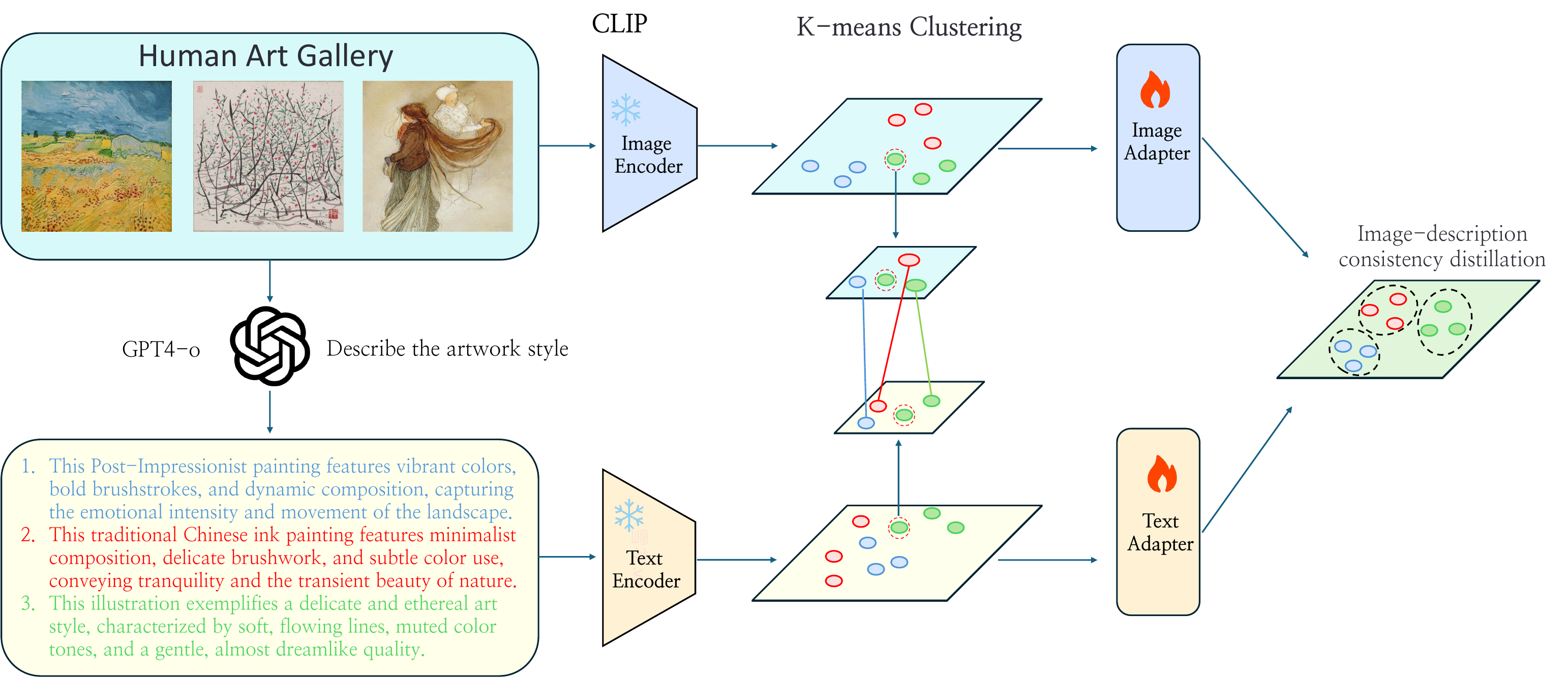}
\end{center}
   \caption{ 
   Description-guided Clustering (DGC) ensures consistent clustering for images with their nearest style neighbors in text space.
}
\label{fig:fig2}
\end{figure*}

\subsection{Legal Practices in Various Countries}
With the rapid development of AIGC technology \cite{gu2022vector,ho2020denoising,dhariwal2021diffusion,ramesh2022hierarchical,podell2023sdxl,esser2024scaling,jia2025secret,jia2023event,huang2025semantic}, countries are actively adjusting laws to address related legal challenges, especially in copyright protection for human works and AI-generated content. In China, the Beijing and Guangzhou Internet Courts have heard cases on "AI-generated text and image copyright infringement" clarifying issues like whether AIGC constitutes a work, its ownership, and potential infringement by AI platforms. In the U.S., Kris Kashtanova's AI-assisted book ``Zarya of the Daw'' \cite{simpson2025auctor} successfully registered copyright, indicating the U.S. Copyright Office's recognition of works predominantly created by humans with AI aid.

Many countries are drafting regulations for AI copyright protection. For instance, China issued the ``Interim Measures for the Management of Generative Artificial Intelligence Services'' \cite{han2010copyright} requiring identification of AI-generated content; Canada passed the ``Artificial Intelligence and Data Act'' \cite{bannerman2023submission} to regulate AI transactions and protect data privacy; the EU emphasizes labeling AI-generated content to reduce misinformation and implements risk-based AI system management \cite{arda2024taxonomy}. These measures reflect the global emphasis on AIGC and its legal framework. Our goal is to establish an intuitive framework to help courts and creators evaluate the copyright status of AI-generated works, including their unique stylistic features and potential infringement.

\subsection{Previous Work on Copyright Protection}
AI-generated content (AIGC) has become widespread due to advancements in AI, particularly in image generation techniques like diffusion models, introducing new copyright challenges. Generative models can replicate features from training datasets, increasing infringement risks.

Various solutions are proposed to address these issues. Gandikota et al. \cite{gandikota2024unified} suggest retraining models to remove certain learned concepts. Kumari et al. \cite{kumari2023ablating} develop "unlearning" techniques to erase memories linked to copyrighted material. Wang et al. \cite{wang2023diagnosis} propose a generative framework to detect and prevent potentially infringing content during creation. Moayeri et al. \cite{moayeri2023text} and Somepalli et al. \cite{somepalli2023diffusion} explore classification methods to summarize artists' stylistic features, aiding in infringement identification.

Researchers also study protection strategies to prevent copying at the source. Wang et al. \cite{wang2023diagnosis} and Cui et al. \cite{cui2025ft} study techniques for adding invisible watermarks to artworks, ensuring traceability even if used to train AI models. Shan et al. \cite{shan2023glaze}, Xue et al. \cite{xue2023toward}, and Zhao et al. \cite{zhao2024can} focus on creating "unlearnable" examples, which models cannot effectively learn even when included in the training set.

Previous work mainly focuses on protecting human artists' unique styles and copyrights. However, whether AI-generated works can possess unique styles and how to protect such works and their creators' intellectual achievements need further exploration. We highlight this gap between AI development and legal practice and propose an effective framework to address this issue.
\section{Method}

\subsection{Motivation}

When examining artworks like human artists, identifying characteristics for a unique artistic style is crucial. Choosing a representative piece per artist as a reference in artwork assessment can provide insights. If artworks can be grouped by an artist's traits and most in a group resemble the central work, it indicates a distinct style.

With the rapid advancement of generative models, the issue of copyright protection for AI-generated artworks has emerged. To address this, we propose a comprehensive evaluation framework for such works and define the following:
$\mathcal{A}=\{A_1,A_2,\dots,A_N\}$ is a set of AI-generated artworks obtained by varying prompt content aspects while keeping stylistic parameters fixed.
$\mathcal{H}=\{H_1,H_2,\dots,H_M\}$ is a collection of artworks created by human artists.
$f_I:\mathcal{X}\to\mathbb{R}^d$ is an embedding function mapping an artwork from the image domain $\mathcal{X}$ to a $d$-dimensional feature space representing visual characteristics.
$f_T:\mathcal{P}\to\mathbb{R}^d$ is an embedding function projecting a prompt from the text domain $\mathcal{P}$ to a $d$-dimensional feature space representing description content. 

We formalize this evaluation as a clustering problem within a feature space that incorporates both stylistic and content attributes of artworks:
\begin{itemize}
    \item \textbf{Consistency of Style Across Multiple Generations}: For any two AI-generated artworks $A_i, A_j \in \mathcal{A}$, the distance between their visual embeddings must be within a small threshold $\epsilon_c$.
   
    \item \textbf{Differentiation from Existing Artists' Styles}: To confirm the distinctiveness of the AI-generated style, for any AI-generated artwork $A_i \in \mathcal{A}$ and any human artwork $H_j \in \mathcal{H}$, the distance between their visual embeddings should surpass a certain threshold $\epsilon_d$.
    
    \item \textbf{Alignment of Content with Prompt Descriptions}: For each AI-generated artwork $A_i$ and its corresponding prompt $P_i$, the similarity between their embeddings should reach or exceed a threshold $\epsilon_a$.
\end{itemize} 

Our objective is to partition the combined dataset $\mathcal{D} = \mathcal{A} \cup \mathcal{H}$ into clusters $\{ C_1, C_2, \dots, C_L \}$. The crucial aspect of this partitioning lies in ensuring that the AI-generated artworks created by a certain artist form a unique and cohesive cluster, which is clearly separated from those artworks representing the styles of human artists as well as other AI artworks with certified copyrights. During partitioning, our aim is to form a specific cluster $C_A$ comprising one's AI-generated artworks. For $C_A$ to be valid and meaningful, it must meet three key criteria, as outlined in the three rules above.

First, the variance of visual embeddings within $C_A$ should not exceed a predefined maximum $\epsilon_c^2$. Mathematically:
\begin{equation}
\frac{1}{|C_A|^2}\sum_{A_i,A_j\in C_A}\|f_I(A_i)-f_I(A_j)\|_2^2\leq\epsilon_c^2 \label{eq1}
\end{equation}
This ensures AI-generated artworks in the cluster have similar visual features, maintaining homogeneity.

Second, to highlight the difference between the AI-generated cluster and those of human artists, we set a separation standard. The minimum distance between any AI-generated artwork in $C_A$ and any human artwork in other clusters $C_H$ should be at least $\epsilon_d$:
\begin{equation}
\min_{A_i\in C_A,H_j\in C_H}\|f_I(A_i)-f_I(H_j)\|_2\geq\epsilon_d \label{eq2}
\end{equation}
This guarantees a clear separation between the two types of artworks based on visual embeddings.

Third, in clustering, we consider both visual (from artworks) and textual (from prompts) features. Let $C_A^I$ be the cluster from image features and $C_A^T$ be the cluster from text features. To ensure reliable and consistent clustering, the Adjusted Mutual Information (AMI) score between $C_A^I$ and $C_A^T$ must meet a certain threshold:
\begin{equation}
\text{AMI}(C_A^I,C_A^T)\geq\epsilon_a \label{eq3}
\end{equation}
Here, $\epsilon_a$ is a high threshold, indicating a strong correlation between visual and text based clustering results. 

In this paper, we set $\epsilon_c=0.60$, $\epsilon_d=0.25$, and $\epsilon_a=0.50$. 

\subsection{Description-Guided Clustering}

As shown in Figure~\ref{fig:fig2}, we propose an approach for clustering artworks by leveraging GPT-4 generated descriptions, which provide rich semantics complementing visual features. Employing cross-modal mutual distillation, we integrate textual and visual modalities to enhance clustering performance. The method consists of feature extraction using CLIP \cite{radford2021learning} encoders, a cross-modal distillation loss, and auxiliary losses for cluster balance and confidence.

Each artwork image $I_i$ is processed by the CLIP visual encoder $E_v$ to obtain visual features, The corresponding description $T_i$ is encoded using the CLIP text encoder $E_t$ to obtain textual features:
\begin{equation}
    v_i = E_v(I_i),    t_i = E_t(T_i).
\end{equation}

Clustering heads $f$ and $g$ produce soft assignments over $K$ clusters:
\begin{equation}
    p_i = f(v_i), \quad q_i = g(t_i), \quad p_i, q_i \in \Delta^K.
\end{equation}

For each sample $i$, we find $\hat{N}$ nearest neighbors $\mathcal{N}_i^v$ and $\mathcal{N}_i^t$ in each modality. The cross-modal distillation loss minimizes discrepancies between modalities:
\begin{equation}
    L_{\text{Dis}} = -\sum_{i} \left( \sum_{j \in \mathcal{N}_i^v} \operatorname{KL}\left( p_j^\tau \| q_i \right) + \sum_{j \in \mathcal{N}_i^t} \operatorname{KL}\left( q_j^\tau \| p_i \right) \right),
\end{equation}
where $p_j^\tau = \operatorname{softmax}\left( p_j / \hat{\tau} \right)$, and $\operatorname{KL}(\cdot\|\cdot)$ denotes the Kullback-Leibler divergence.

To encourage confident and balanced clustering, we define the confidence loss:
\begin{equation}
    L_{\text{Con}} = -\sum_{i} \log \left( p_i^\top q_i \right),
\end{equation}
and the cluster entropy loss:
\begin{equation}
    L_{\text{Ent}} = -\sum_{k} \left( \bar{p}_k \log \bar{p}_k + \bar{q}_k \log \bar{q}_k \right),
\end{equation}
where $\bar{p}_k = \tfrac{1}{n} \sum_{i} p_{ik}$ and $\bar{q}_k = \tfrac{1}{n} \sum_{i} q_{ik}$.

The total loss is:
\begin{equation}
    L_{\text{Total}} = L_{\text{Dis}} + L_{\text{Con}} - \alpha L_{\text{Ent}},
\end{equation}
with hyperparameter $\alpha > 0$.

In experiments, we set $\hat{N} = 10$, $\alpha = 3$, and $\hat{\tau} = 0.1$. Features are normalized to unit norm. 

By integrating semantic descriptions with visual features, our method enhances discriminative power, achieving accurate clustering. Cross-modal distillation aligns assignments across modalities, and balance loss prevents cluster collapse, ensuring meaningful clusters. This framework adapts to other multimodal tasks, aiding in organizing complex datasets.

\begin{figure*}[h!]
\begin{center}
\includegraphics[width=0.9\linewidth]{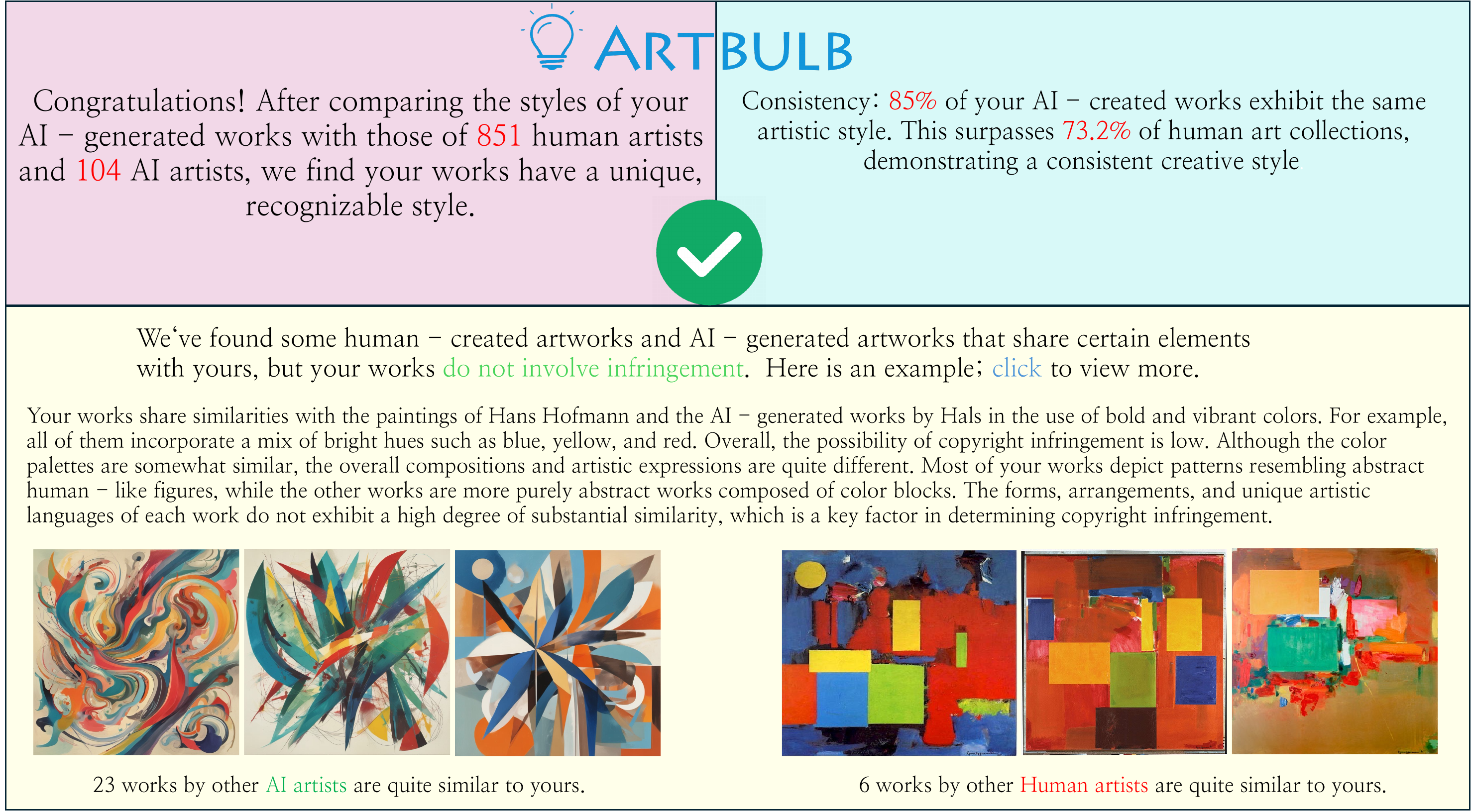}
\end{center}
   \caption{ Our main contribution is the creation of ArtBulb, an interpretable and quantifiable framework based on legal research. ArtBulb uses clustering to address style infringement issues, determining if AI-generated works have a unique style based on three key criteria. It also produces a clear, quantifiable report summarizing its findings.}
\label{fig:fig3}
\end{figure*}

\subsection{Collection of the AI Copyright Dataset (AICD)}

\subsubsection{Building the Reference Corpus}

To analyze human art styles and spot AI-generated content, we create a comprehensive reference corpus as part of the AICD. This corpus assesses artists' unique styles and forms clusters with clear authorship in our framework.

We pick works from 567 WikiArt \cite{artgan2018} artists, each having at least 30 pieces, for diversity. Aware of WikiArt's limitations, we expand the dataset with other sources: 672 film and game illustrators, 593 children's picture book artists, 781 contemporary artists, and 172 Chinese painters. These art forms differ from WikiArt's traditional paintings, enriching the corpus. Each new source artist also has over 30 works for consistency.

We use our clustering algorithm on the combined data to analyze human art styles. After clustering artists with similar styles, we screen the clusters and keep those where over 50\% of works are from one artist with clear copyright. Eventually, we obtain 364,297 artworks from 2,785 artists with distinct personal styles. This corpus helps determine if AI-generated works have unique styles meeting legal criteria.

\subsubsection{Creating the AICD Evaluation Dataset}

To tackle the shortage of datasets for assessing the copyright of AI-generated artworks, we construct the AICD evaluation dataset, which serves as a precious resource for this field. We gather 1,786 AI-generated works with identified creators and detailed prompts from multiple sources, such as social media, AI art demonstrations at conferences, and online gallery exhibitions. Additionally, we adopt the synthetic method from CopyCat to generate a dataset for potential intellectual property infringement cases.

Five independent artists and legal experts in copyright evaluate each pair of images. They assign scores ranging from 1 to 5, where higher scores indicate greater confidence in independent copyright. We retain images with an average score above 4.0. As a result, 542 AI-generated artworks are regarded as having copyrights.

To incorporate AI-generated works without independent copyrights, we use common prompts like ``generate an image of a dog'' to create 2,000 images. Furthermore, by referring to works with clear human copyrights, we generate 5,000 AI images that may potentially infringe on these copyrights. After screening by art and copyright law professionals, we select 1,278 images with clear infringement. These are added to the validation set as negative samples. By combining these negative samples with the positive examples, we form an evaluation dataset to test the accuracy of our proposed framework and existing MLLMs.

The assembled AICD evaluation dataset, covering AI-generated works with/without independent copyrights and potential infringement cases, highlights our contribution of providing valuable resources for future research.

\begin{figure*}[h!]
\begin{center}
\includegraphics[width=0.82\linewidth]{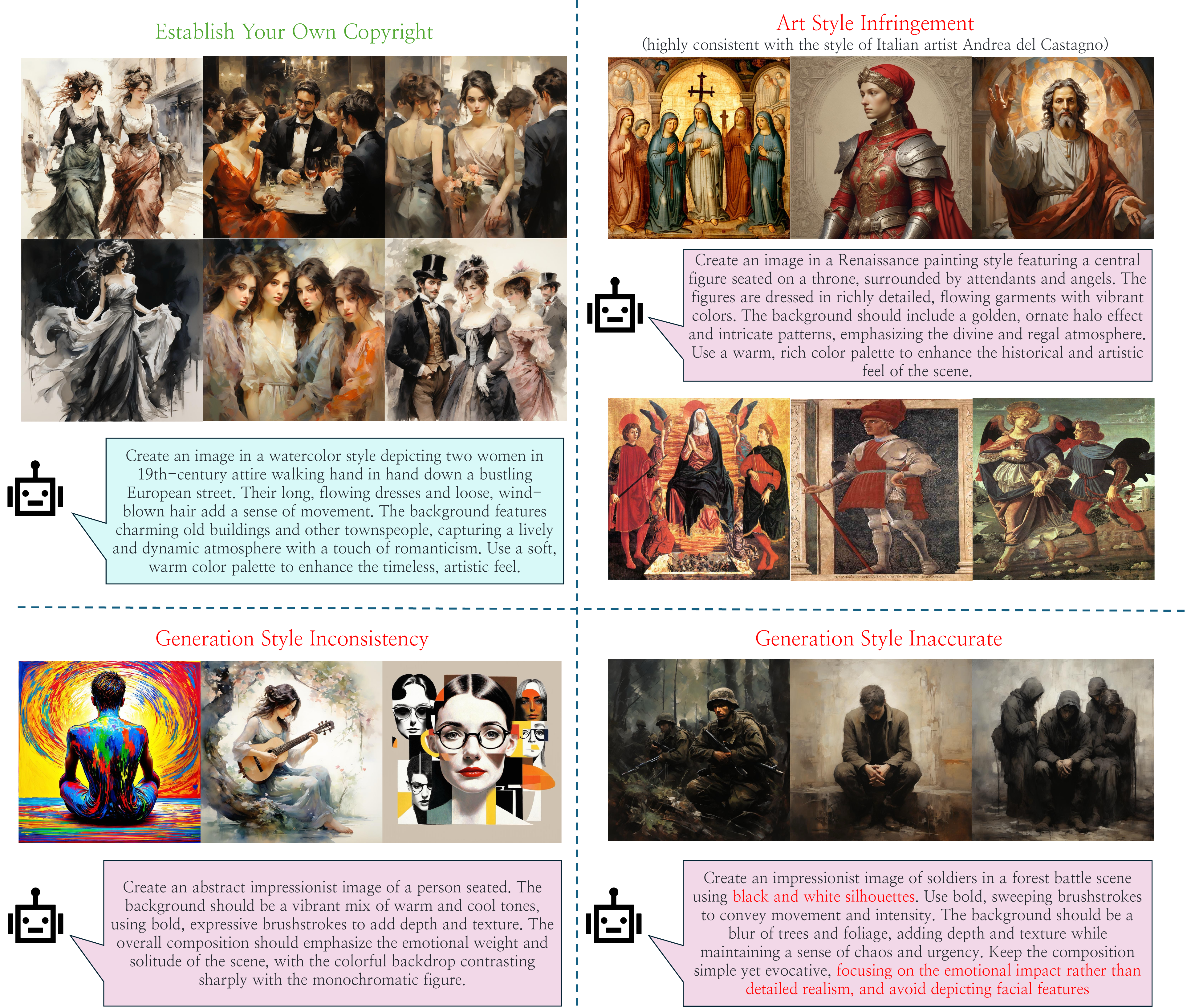}
\end{center}
   \caption{Case study of four different judgment outcomes. ArtBulb can accurately determine whether AI-generated works should be protected or not recognized for certain reasons. Protected works will be included and the copyright dataset will be dynamically updated.
}
\label{fig:fig4}
\end{figure*}

\subsection{Explainable Copyright Framework: ArtBulb}

\textbf{ArtBulb} is designed to address practical needs by helping generative AI artists determine if their works exhibit a distinct personal style. This capability is critical for assessing potential copyright infringement of human artworks and establishing authorship rights for AI-generated content.

The workflow begins by augmenting input prompts with entity substitution to generate additional samples using the same model, expanding the dataset for robust evaluation. These augmented outputs are then processed through a feature extraction network. A representative artwork is selected as a cluster centroid and grouped with existing human art clusters to analyze stylistic proximity between AI and human works. Finally, clustering results are interpreted via a chain-of-thought reasoning process using a multimodal large language model (MLLM). In our implementation, GPT-4o generates a detailed and actionable AI copyright detection report by cross-analyzing AI cluster characteristics against the most similar human art clusters.

Figure~\ref{fig:fig3} presents a detailed report generated by ArtBulb for a real AI artist concerned with the originality of their work. The artist provided five stylistically similar pieces, along with the corresponding prompts and the generative model used. ArtBulb produced a clear and interpretable report to assess whether these works establish their copyright.

A key innovation is ArtBulb's iterative database enhancement. When AI-generated works are validated as unique, they are integrated into the existing database as new copyright clusters. This continuous expansion improves reference datasets for future evaluations, enhancing detection accuracy over time.

\section{Experiments}

\subsection{Setup}
\subsubsection{Evaluation Metrics}

To evaluate the clustering performance, we used three widely adopted metrics: Normalized Mutual Information (NMI), Accuracy (ACC), and the Adjusted Rand Index (ARI). Higher values of these metrics indicate better clustering results. To determine whether AI-generated works have their own copyright, we compared them with annotations made by human experts. Specifically, we measured the consistency between the AI outputs and the expert annotations using binary classification accuracy and F1-score.

\subsubsection{Implementation Details}

We fine-tuned the clustering adapter model for 10 epochs to align the text descriptions of images with their styles in the clustering results. To ensure a fair comparison, we also fine-tuned the CSD \cite{somepalli2024measuring} and ARTSAVANT \cite{moayerirethinking} models using the same data as the clustering training. The initial learning rate was set to $1 \times 10^{-4}$, and a cosine annealing learning rate decay strategy was applied throughout all training processes. We used the AdamW optimizer \cite{loshchilov2017decoupled} for training. Additionally, GPT-4o generated the descriptive information for all artworks. All experiments were conducted on 4 NVIDIA A100 GPUs.

\subsection{Performance}

\begin{table*}[h!]
    \centering
    \caption{The clustering results of the styles of real artists in the AICD dataset}
    \resizebox{\textwidth}{!}{
    \begin{tabular}{l|cccccccccccccccccccc}
        \toprule
        \textbf{Dataset} & \multicolumn{3}{c}{\textbf{WikiART}} & \multicolumn{3}{c}{\textbf{Video Game}} & \multicolumn{3}{c}{\textbf{Contemporary Art}} & \multicolumn{3}{c}{\textbf{Chinese Art}} & \multicolumn{3}{c}{\textbf{Children Book}} \\
        \cmidrule(lr){2-4} \cmidrule(lr){5-7} \cmidrule(lr){8-10} \cmidrule(lr){11-13} \cmidrule(lr){14-16}
        & \textbf{ACC} & \textbf{NMI} & \textbf{ARI} & \textbf{ACC} & \textbf{NMI} & \textbf{ARI} & \textbf{ACC} & \textbf{NMI} & \textbf{ARI} & \textbf{ACC} & \textbf{NMI} & \textbf{ARI} & \textbf{ACC} & \textbf{NMI} & \textbf{ARI} \\
        \midrule
         DEC \cite{xie2016unsupervised} & 10.15 & 30.42 & 5.73 & 12.37 & 25.69 & 3.24 & 8.59 & 20.85 & 2.37 & 10.81 & 22.58 & 1.69 & 9.95 & 21.47 & 2.87 \\ 
         DAC \cite{chang2017deep} & 12.48 & 35.65 & 6.87 & 14.26 & 30.15 & 4.58 & 10.91 & 22.76 & 3.42 & 12.53 & 25.33 & 2.76 & 11.65 & 24.57 & 3.10 \\
         SCAN \cite{van2020scan} & 34.87 & 50.21 & 11.92 & 36.76 & 45.85 & 5.64 & 22.32 & 34.65 & 4.25 & \underline{34.58} & 38.15 & \underline{3.87} & 23.75 & 36.47 & 3.65 \\
         SIC \cite{cai2023semantic} & \underline{36.43} & \underline{59.13} & 14.36 & \underline{38.53} & \underline{53.76} & \underline{6.87} & 34.15 & 46.87 & 5.14 & 34.96 & 40.48 & \textbf{4.08} & \underline{35.86} & \underline{44.37} & \underline{5.76} \\
        \midrule
        MOCOv3 \cite{chen2021empirical} (k-means) & 27.20 & 49.80 & 7.39 & 31.45 & 41.30 & 2.98 & 26.10 & 38.57 & 1.03 & 27.43 & 35.85 & 1.44 & 28.81 & 40.59 & 4.54 \\
        DINOv2 \cite{oquab2023dinov2} (k-means) & 28.63 & 41.63 & 2.54 & 33.29 & 47.68 & 5.32 & 27.08 & 43.12 & 2.04 & 33.08 & 39.85 & 2.29 & 29.99 & 39.42 & 4.20 \\
        CLIP \cite{radford2021learning} (k-means) & 36.09 & 58.98 & \underline{14.85} & 35.91 & 53.42 & 6.17 & 34.13 & 49.56 & \underline{6.02} & 33.59 & 42.86 & 3.54 & 33.91 & 42.34 & 4.86 \\
        BLIP2 \cite{li2023blip} (k-means) & 34.36 & 58.21 & 13.90 & 37.31 & 52.93 & 6.27 & \underline{36.05} & \underline{50.12} & 5.53 & \underline{34.58} & \underline{43.25} & 3.76 & 34.03 & 42.51 & 5.07 \\
        \midrule
        DGC (Ours) & \textbf{41.45} & \textbf{61.20} & \textbf{15.72} & \textbf{43.49} & \textbf{55.74} & \textbf{7.88} & \textbf{42.84} & \textbf{55.39} & \textbf{7.04} & \textbf{39.07} & \textbf{44.31} & \underline{4.01} & \textbf{40.83} & \textbf{46.72} & \textbf{6.04} \\
        \bottomrule
    \end{tabular}
    }
    \label{tab:table1}
\end{table*}

\begin{table*}[htbp]
    \centering
    \caption{The accuracy rates and F1 scores of different models on the AICD dataset}
    \resizebox{\textwidth}{!}{
    \begin{tabular}{l|*{14}{c}}
        \toprule
        \textbf{Method} & \multicolumn{2}{c}{\makecell{\textbf{Western Art}}}& \multicolumn{2}{c}{\makecell{\textbf{Chinese Art}}}& \multicolumn{2}{c}{\makecell{\textbf{Comic}\\ \textbf{Line-Drawing}}}& \multicolumn{2}{c}{\makecell{\textbf{Video Game}}}& \multicolumn{2}{c}{\makecell{\textbf{Children Book}}}& \multicolumn{2}{c}{\makecell{\textbf{Graphic Design}}}& \multicolumn{2}{c}{\makecell{\textbf{Average}}}\\
        \cmidrule(r){2-3} \cmidrule(r){4-5} \cmidrule(r){6-7} \cmidrule(r){8-9} \cmidrule(r){10-11} \cmidrule(r){12-13} \cmidrule(r){14-15}
        & \makecell{\textbf{ACC}} & \makecell{\textbf{F1}} & \makecell{\textbf{ACC}} & \makecell{\textbf{F1}} & \makecell{\textbf{ACC}} & \makecell{\textbf{F1}} & \makecell{\textbf{ACC}} & \makecell{\textbf{F1}} & \makecell{\textbf{ACC}} & \makecell{\textbf{F1}} & \makecell{\textbf{ACC}} & \makecell{\textbf{F1}} & \makecell{\textbf{ACC}} & \makecell{\textbf{F1}} \\
        \midrule
        GPT-4o \cite{achiam2023gpt}  & 0.35 & 0.46 & 0.25 & 0.09&  0.52 & 0.58 & 0.49 & 0.44 & 0.50 & 0.47 & 0.43 & 0.43 & 0.42 & 0.41 \\
        Qwen2-VL \cite{wang2024qwen2} & 0.30 & 0.40 & 0.07 & 0.23 & 0.40 & 0.42 & 0.46 & 0.45 & 0.45 & 0.45 & 0.37 & 0.32 & 0.34 & 0.38 \\
        DeepSeek-VL \cite{lu2024deepseek}  & 0.27 & 0.40 & 0.11 & 0.09 & 0.33 & 0.46 & 0.49 & 0.46 & 0.41 & 0.43 & 0.41 & 0.34 & 0.34 & 0.36\\
        CSD \cite{somepalli2024measuring} & 0.82 & 0.83 & \underline{0.80} & 0.77 & 0.79 & 0.80 & 0.73 & 0.77 & 0.76 & 0.82 & 0.75 & 0.74 & 0.77 & 0.79 \\
        ARTSAVANT \cite{moayerirethinking} & \textbf{0.86} & \underline{0.84} & 0.75 & 0.75 & \underline{0.84} & \underline{0.83} & 0.75 & 0.77 & \underline{0.86} & \underline{0.84} & 0.74 & \underline{0.81} & 0.80 & \underline{0.80} \\
        \midrule
        DGC & 0.84 & 0.83 & \textbf{0.87} & \textbf{0.88} & 0.81 & 0.77 & \underline{0.80} & \underline{0.82} & 0.76 & 0.70 & \underline{0.78} & 0.75 & \underline{0.81} & 0.79 \\
        DGC w/ ArtBulb & \underline{0.85} & \textbf{0.85} & \textbf{0.87} & \underline{0.86} & \textbf{0.88} & \textbf{0.89} & \textbf{0.87} & \textbf{0.88} & \textbf{0.87} & \textbf{0.88} & \textbf{0.83} & \textbf{0.83} & \textbf{0.86} & \textbf{0.88} \\
        \bottomrule
    \end{tabular}
    \label{tab:table2}
    }
\end{table*}

\subsubsection{Performance of Artistic Styles Clustering}

We assess our proposed Description Guided Clustering (DGC) method on the WikiART dataset and four other art style datasets and compare it with deep clustering baseline methods. Table~\ref{tab:table1} shows that DGC can effectively capture interpretable image style descriptions. With our cross modal mutual distillation strategy, DGC achieves state of the art clustering performance and significantly outperforms previous methods on all five datasets.

Notably, art style clustering is much more challenging than category based clustering on datasets such as ImageNet. This is because the styles of many artists overlap due to factors like historical periods, art movements, and the influence of predecessors. These results prove that our simple yet effective strategy fully exploits the capabilities of visual language models in art style clustering and outperforms the traditional approach of directly clustering pretrained features.

\begin{table}[h]
    \centering
    \small 
    \caption{Ablation studies on the varying $\alpha$ and $\hat{N}$.}
    \label{tab:ablation_study}
    \begin{tabular*}{0.47\textwidth}{@{\extracolsep{\fill}} c ccc c ccc}
        \toprule
        & \multicolumn{3}{c}{$\alpha$} & \quad & \multicolumn{3}{c}{$\hat{N}$} \\
        \cmidrule(lr){2-4} \cmidrule(lr){6-8}
        Value & 2 & 3 & 5 & & 5 & 10 & 15 \\
        \midrule
        ACC & 35.38 & 41.45 & 40.29 & & 39.78 & 41.45 & 40.73 \\
        NMI & 53.94 & 61.20 & 58.72 & & 56.86 & 61.20 & 60.54 \\
        ARI & 9.46 & 15.72 & 12.33 & & 14.38 & 15.72 & 15.02 \\
        \bottomrule
    \end{tabular*}
\end{table}

\begin{table}[h]
    \centering
    \small
    \caption{Ablation study of hyperparameters $\epsilon_c$ (Consistency), $\epsilon_d$ (Differentiation), and $\epsilon_a$ (Alignment) for result judgment.}
    \label{tab:ablation_epsilon}
    \begin{tabular*}{0.4\textwidth}{@{\extracolsep{\fill}} c c c c c}
        \toprule
        Parameter & Value & Precision & Recall & F1-score \\
        \midrule
        $\epsilon_c$ & 0.55 & 0.91 & 0.72 & 0.80 \\
                     & 0.60 & 0.88 & 0.87 & \textbf{0.88} \\
                     & 0.65 & 0.79 & 0.91 & 0.85 \\
        \midrule
        $\epsilon_d$ & 0.20 & 0.75 & 0.89 & 0.81 \\
                     & 0.25 & 0.88 & 0.87 & \textbf{0.88} \\
                     & 0.30 & 0.90 & 0.75 & 0.82 \\
        \midrule
        $\epsilon_a$ & 0.45 & 0.83 & 0.88 & 0.85 \\
                     & 0.50 & 0.88 & 0.87 & \textbf{0.88} \\
                     & 0.55 & 0.89 & 0.82 & 0.85 \\
        \bottomrule
    \end{tabular*}
\end{table}

\subsubsection{Binary Classification of AI Copyrighted Works}
To assess the capabilities of various models in determining potential copyright issues of AI generated artworks, a simple binary classification task is initiated. This task quantifies each model's discriminative ability based on annotations from multiple human experts. As shown in Table~\ref{tab:table2}, when using only vision language models to evaluate image pairs, even the top performing model, GPT-4o, has an accuracy of only 42\%.

In contrast to classification based copyright judgment frameworks such as CSD and ARTSAVANT, the proposed clustering based method, DGC, shows significantly higher accuracy in identifying copyright concerns. Notably, classification approaches are not suitable for the task of copyright assessment in AI generated works. This is because such frameworks cannot handle new AI art styles without retraining and are unable to incorporate AI artworks with confirmed copyrights into a protection repository.

The clustering method effectively addresses these limitations. Moreover, instead of relying solely on clustering results for copyright attribution, accuracy is enhanced and clear explanations are provided by integrating clustering outcomes with MLLMs. This integration is particularly important in the field of copyright protection, where transparency and interpretability are crucial.

\subsubsection{Quality Assessment by Legal Professionals} 

Since copyright determination requires interpretability and evaluating explanations presents challenges with objective metrics, we adopt the Average Human Ranking (AHR) \cite{song2024preference} as a preference measurement. In this metric, copyright protection experts rate explanations on a scale from 1 to 5, with higher scores indicating better quality. Ten legal experts and artists assess the explanation quality of copyright judgments for a randomly selected set of 100 AI-generated works in AICD (50\% involving copyright and 50\% not). Guiding models with DGC clustering results significantly improves explanation quality scores: Qwen2-VL's score increases from 1.80 to 3.55, and GPT-4o's score improves from 2.05 to 3.70, aligning with evaluator expectations. 

Additionally, ArtBulb undergoes testing with 19 infringing images from real legal cases. Five non-technical legal professionals evaluate it, giving an average score of 4.50 out of 5 across four criteria: user-friendly interface, clear language, detailed analysis, and actionable recommendations. Supplementary material contains further details.

\subsection{Case Study}

Figure~\ref{fig:fig4} illustrates 4 types of copyright judgments made by ArtBulm on different AI-artworks: one affirming copyright and three denying it. We observe that, unlike general-purpose MLLMs, ArtBulm emphasizes artistic style over mere content comparison. It effectively mimics human judgment, providing convincing reasons for its decisions, and adapts well to evaluating works across various styles.

\subsection{Ablation Study}

To validate the effectiveness of each component in ArtBulb, we conduct ablation studies focusing on feature selection for clustering, parameter choices in the clustering model, and the selection of MLLMs.

\textbf{Feature Selection}: We compare the impact of different features on clustering accuracy. Tests on the WikiART dataset reveal that replacing CLIP features with BLIP-2 leads to a 1.78\% drop in accuracy (ACC), while replacing them with SigLip results in a 2.54\% decrease. 

\textbf{Clustering Model Parameter Selection}: We investigate how different parameter settings affect clustering results. As shown in Table~\ref{tab:ablation_study}, the WikiART tests indicate that selecting an optimal number of neighbor samples enhances clustering accuracy.  We also analyze how different settings of the hyperparameters $\epsilon_c$ (Consistency), $\epsilon_d$ (Differentiation), and $\epsilon_a$ (Alignment) affect the ArtBulb performance. As shown in Table~\ref{tab:ablation_epsilon}, the optimal F1-score of 0.88 is consistently achieved when $\epsilon_c=0.60$, $\epsilon_d=0.25$, and $\epsilon_a=0.50$. 

\textbf{MLLMs Selection}: We evaluate GPT4-o, Qwen2-VL, and DeepSeek-VL using the same prompts. On the \textit{AICD} test set, GPT4-o achieves an average accuracy of 0.86, outperforming Qwen2-VL and DeepSeek-VL, which score 0.81 and 0.74 respectively.

\section{Conclusion}
This paper investigates whether AI-generated art can have a unique style deserving of legal copyright protection. Analyzing legal precedents, we define three criteria for independent artistic style: consistency, uniqueness, and accuracy. We introduce ArtBulb, an interpretable framework for AI art copyright judgment that combines a novel style description-based multimodal clustering method with MLLMs. We also present AICD, the first benchmark dataset for AI art copyright. Our work bridges the legal and technological communities and draws attention to AI art copyright issues.
{
    \small
    \bibliographystyle{ieeenat_fullname}
    \bibliography{main}
}

\end{document}